\documentclass[letterpaper, 10 pt, conference]{ieeeconf}  

\IEEEoverridecommandlockouts                              

\usepackage{graphicx}
\usepackage{xcolor}
\usepackage{color, colortbl}
\usepackage{multirow}
\definecolor{Gray}{gray}{0.9}
\definecolor{LBlue}{rgb}{0,0.2,0.8}
\definecolor{DGreen}{rgb}{0,0.59,0}
\definecolor{LGreen}{rgb}{0.89,1,0.89}
\definecolor{Gray2}{gray}{0.3}
\usepackage{amsfonts}

\title{\LARGE \bf
AOP-Net: \underline{A}ll-in-\underline{O}ne \underline{P}erception \underline{Net}work for Joint LiDAR-based 3D Object Detection and Panoptic Segmentation
}

\author{Yixuan Xu$^{*}$, Hamidreza Fazlali$^{*}$, Yuan Ren, and Bingbing Liu
\\
  Huawei Noah's Ark Lab \\
  \texttt{\{richard.xu2, hamidreza.fazlali1, yuan.ren3, liu.bingbing\}@huawei.com} 
\thanks{$^{*}$ indicates equal contribution.}%
}

\begin{document}

\maketitle
\thispagestyle{empty}
\pagestyle{empty}

\begin{abstract}

LiDAR-based 3D object detection and panoptic segmentation are two crucial tasks in the perception systems of autonomous vehicles and robots. In this paper, we propose All-in-One Perception Network (AOP-Net), a LiDAR-based multi-task framework that combines 3D object detection and panoptic segmentation. In this method, a dual-task 3D backbone is developed to extract both panoptic- and detection-level features from the input LiDAR point cloud. Also, a new 2D backbone that intertwines Multi-Layer Perceptron (MLP) and convolution layers is designed to further improve the detection task performance. Finally, a novel module is proposed to guide the detection head by recovering useful features discarded during down-sampling operations in the 3D backbone. This module leverages estimated instance segmentation masks to recover detailed information from each candidate object. The AOP-Net achieves state-of-the-art performance for published works on the nuScenes benchmark for both 3D object detection and panoptic segmentation tasks. Also, experiments show that our method easily adapts to and significantly improves the performance of any BEV-based 3D object detection method.

\end{abstract}

\section{INTRODUCTION}
Understanding the surrounding 3D environment is an essential component in autonomous driving and robotics to ensure safety and reliability. LiDAR-based 3D object detection and panoptic segmentation are two common tasks performed by the perception systems. For 3D object detection, foreground objects such as cars, pedestrians, etc., are classified and localized by 3D bounding boxes.
For 3D panoptic segmentation, each point in the scene is categorized with a semantic label and points for the same foreground object are assigned a unique instance ID.
For efficiency, most detection methods \cite{zhou2018voxelnet, lang2019pointpillars, yang2018pixor} attempt to extract features from a summarized representation of the scene. Some quantize LiDAR points into volumetric grids, known as voxels, and then process the voxels with a 3D Convolutional Neural Network (CNN). Others project the point cloud or 3D voxels into 2D grids in Bird's-Eye-View (BEV) or Range-View (RV) and process the grids by a 2D CNN. Furthermore, the CNNs deployed
typically perform 
down-sampling steps to enlarge the receptive fields of convolution kernels and extract features efficiently. However, while quantization, projection, and down-sampling reduce computational cost, they result in considerable information loss about the scene.

Likewise, LiDAR-based 3D panoptic segmentation methods \cite{razani2021gp, li2021cpseg, milioto2020lidar, hurtado2020mopt} follow similar point cloud data representation strategies. While recent 3D object detection methods mostly operate in the scale-invariant BEV plane
\cite{lang2019pointpillars, yin2021center, chen2020object}, many 3D panoptic segmentation methods rely on the denser and more detailed object representations in RV
\cite{razani2021gp, li2021cpseg, sirohi2021efficientlps}. Considering the strengths of each projection view and complementary goals of each perception task, \cite{cvpr} demonstrates that information extracted by the backbone of RV-based panoptic segmentation model can also be helpful for object detection. This approach presents a question: can object detection and panoptic segmentation networks be more integrated, so that both tasks benefit from one another?

To this end, we propose the All-in-One Perception Network (AOP-Net) for LiDAR-based joint 3D object detection and panoptic segmentation. In this multi-task framework, 3D object detection and panoptic segmentation are jointly trained and take advantage of one another for performance gains. More specifically, a dual-task 3D backbone is developed to extract both detection- and panoptic-level features from the voxelized 3D space. A new 2D backbone for 3D object detection is proposed that extensively fuses Multi-Layer Perceptron (MLP) layers into CNN, enabling a larger receptive field and deeper pixel-wise feature extraction while exhibiting a similar model complexity compared to traditional 2D backbones used for detection \cite{lang2019pointpillars, yin2021center}. Finally, to recover lost useful features due to down-sampling, a novel Instance-based Feature Retrieval (IFR) module is proposed, which leverages the instance-level estimation from panoptic segmentation to recover object-specific features and highlight corresponding locations to guide object detection. 
Our contributions can be summarized into four-fold: 1) A multi-task framework is proposed for joint LiDAR-based 3D object detection and panoptic segmentation. In this method, both tasks achieve performance gains as they mutually benefit from one another.
2) A deep and efficient 2D backbone that mixes MLPs and convolution layers for 3D object detection.
3) The IFR module that augments the detection head and recovers useful discarded multi-scale features based on panoptic segmentation estimations.
4) Through experiments, we show that each new component provides effective performance gain, and that the proposed framework easily adapts to and improves the performance of any BEV-based 3D object detection method.

\section{Related Work}

\subsection{3D Object Detection}

Efficient 3D object detection methods quantize the 3D space using small voxel grids and operate on the BEV plane. Then, features are extracted to encode each voxel. VoxelNet \cite{zhou2018voxelnet} designs a learnable Voxel Feature Encoder (VFE) layer to encode points inside each voxel and then exploits a 3D CNN to extract features across voxel grids. SECOND \cite{yan2018second} proposes 3D Sparse convolution layers to reduce 
the computations of 3D convolution by leveraging the sparsity of voxel grids.
PointPillars \cite{lang2019pointpillars} further improves the inference speed by reducing the voxel number along the height dimension to one and using a 
2D CNN to process the generated pseudo image. 
CenterPoint \cite{yin2021center} is an anchor-free object detection method that addresses the challenge caused by anchor-based methods.
CenterPoint designed a center-based detection head for detecting the center of 3D boxes in BEV plane. This approach significantly improves the detection accuracy as it does not need to fit axis-aligned boxes to rotated objects.

\subsection{3D Panoptic Segmentation}
3D panoptic segmentation methods usually extend from an RV-based semantic segmentation network, with an additional mechanism that groups foreground points into clusters, each representing a segmented instance. LPSAD \cite{milioto2020lidar} uses a shared encoder with two decoders, where the first decoder predicts semantic tags and the second predicts the center offset for each foreground point, and subsequently it uses an external algorithm such as BFS and HDBSCAN \cite{campello_hdbscan} to group nearby shifted points into the same cluster.
Panoster \cite{gasperini2021panoster} uses a learnable clustering method to assign instance labels to each point.
CPSeg \cite{li2021cpseg} is a cluster-free panoptic segmentation method that segments objects by pillarizing points according to their learned embeddings and finding connected pillars through a pairwise embedding comparison.

\subsection{3D Multi-task Perception}
Few attempts have been made to leverage the complementary nature of segmentation and detection tasks. PointPainting \cite{vora2020pointpainting} and FusionPainting \cite{fusionPainting} append semantic class scores from pretrained segmentation networks to the point cloud before feeding to a 3D object detection model. A similar method \cite{cvpr} to our framework was introduced recently, in which a panoptic segmentation model and an object detection model are jointly trained. Its Cascade Feature Fusion Module fuses BEV and RV features from detection and panoptic segmentation backbone, respectively. Its class-wise foreground attention module embeds predicted foreground semantic scores in detection features. In \cite{cvpr}, although panoptic segmentation is leveraged to bring improvement to object detection, the two tasks fail to mutually benefit.


\begin{figure*}[t]
    \centering
    \includegraphics[width=1\linewidth]{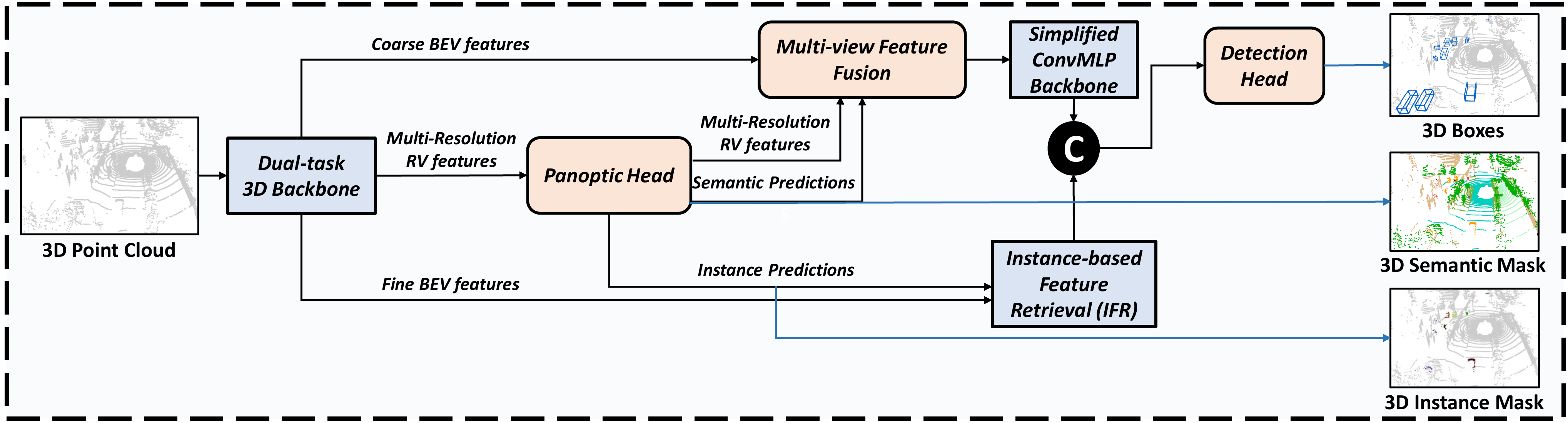}
    \caption{Overall framework of the proposed joint 3D object detection and panoptic segmentation. The proposed modules are shown with \textcolor{blue}{blue} color. Best viewed in color.}
    \label{fig:model}
\end{figure*}

\section{Method}
\subsection{Overview}

We propose a framework that jointly performs 3D object detection and panoptic segmentation 
as shown in Figure \ref{fig:model}. In this multi-task method, a BEV-based 3D object detection model and an RV-based 3D panoptic segmentation model are deeply integrated, so that the performance of both tasks can improve substantially. We exploit a simplified version of CPSeg \cite{li2021cpseg}, a U-Net architecture with two task-specific decoders, for panoptic segmentation due to its real-time performance and high accuracy.
For object detection, we rely on the detection head from the CenterPoint \cite{yin2021center} for its superior performance.

To integrate the two tasks into one unified framework, we propose a dual-task 3D backbone to extract multi-scale features from voxelized point cloud. 
These features are compressed and projected to the RV plane, fused with the set of features extracted directly from the RV-projected point cloud via three Convolutional Bottleneck Attention Modules (CBAM) \cite{woo2018cbam}, and fed to the panoptic head.
This lightweight operation effectively augments the panoptic head with detection-level features. To introduce panoptic-level features to object detection, we exploit the cascade feature fusion and class-wise foreground attention modules in \cite{cvpr}, shown as Multi-view Feature Fusion in Figure \ref{fig:model}. 

The lowest resolution voxel features from the dual-task 3D backbone are projected to BEV for the object detection task. 
These features encode the instance- and semantic-level information besides the detection-level information.
Also, inspired by \cite{li2021convmlp}, we propose a more effective 2D backbone that mixes MLPs with convolutional layers to process the features for the detection head. Moreover, a novel IFR module augments the detection head by leveraging the predicted instance masks to recover relevant object features that are otherwise lost during down-sampling operations in the dual-task 3D backbone.
Details of the proposed modules are described below.

\begin{figure*}[t]
    \centering
    \includegraphics[width=1\linewidth]{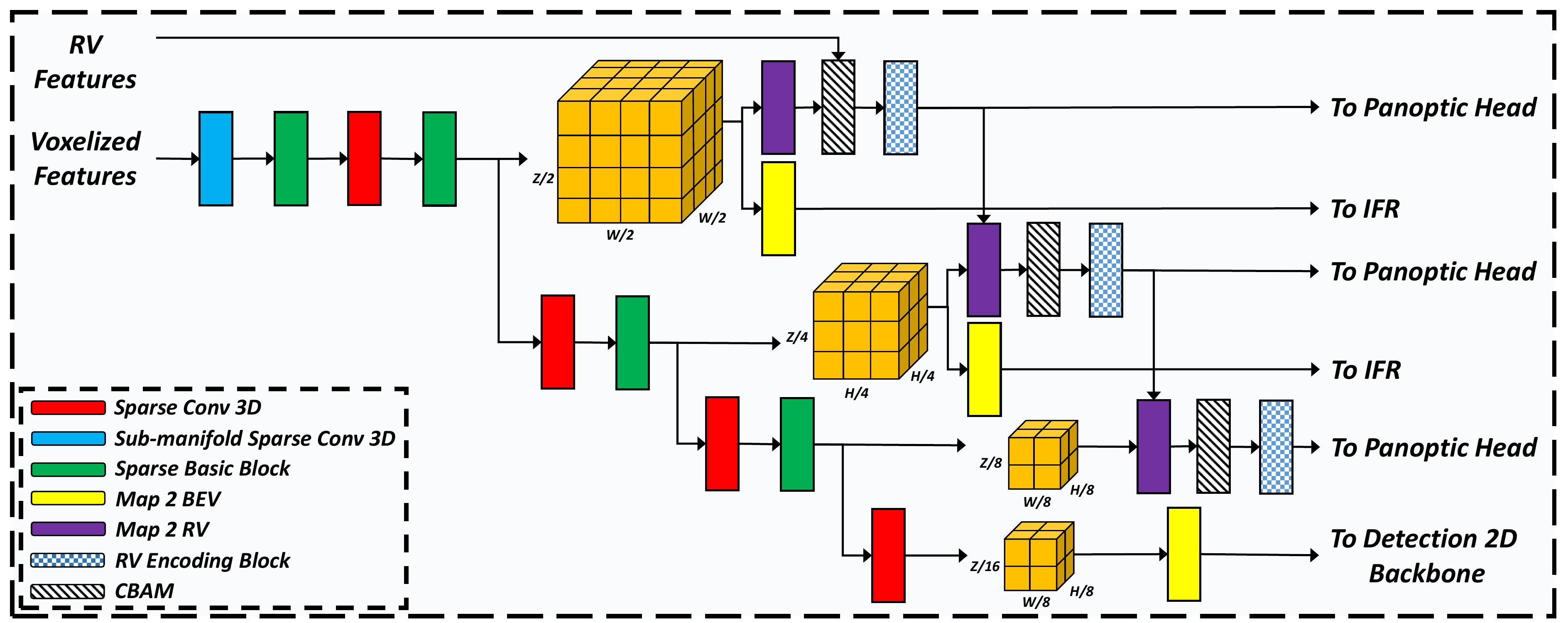}
    \caption{Architecture of the dual-task 3D backbone in the proposed multi-task framework. Best viewed in color.}
    \label{fig:3dbackbone}
\end{figure*}

\subsection{Dual-task 3D Backbone}
Shown in Figure \ref{fig:3dbackbone}, the 3D backbone exploited in our method 
is responsible for extracting features from 3D voxels. 

To efficiently transfer features from 3D backbone for the object detection task, we follow \cite{zhou2018voxelnet, yan2018second, yin2021center} and map 3D features in the coarsest resolution $(\frac{Z}{16} \times \frac{H}{8} \times \frac{W}{8})$ to BEV and feed them to the 2D backbone. However, in contrast to former methods,
detailed object information embedded in two sets of higher resolution voxel features will be recovered later in the IFR module. Moreover, three sets of higher resolution voxel features are projected to RV, fused with features extracted directly from the RV-projected point cloud via corresponding CBAMs, and processed by CPSeg's RV encoding blocks. These multi-scale voxel-based features augment the RV-based panoptic head.
Meanwhile, this augmentation also enforces the 3D backbone to develop a richer set of semantic- and instance-level features.

\subsection{Simplified ConvMLP (SC) Backbone}
Recently, MLP-based vision backbones are 
receiving more attention \cite{tolstikhin2021mlp, touvron2021resmlp, melas2021you, chen2021cyclemlp, li2021convmlp} for their ability to compete or even perform better than fully convolution-based backbones in dense vision prediction tasks.

\begin{figure}[t]
    \centering
    \includegraphics[width=1\linewidth]{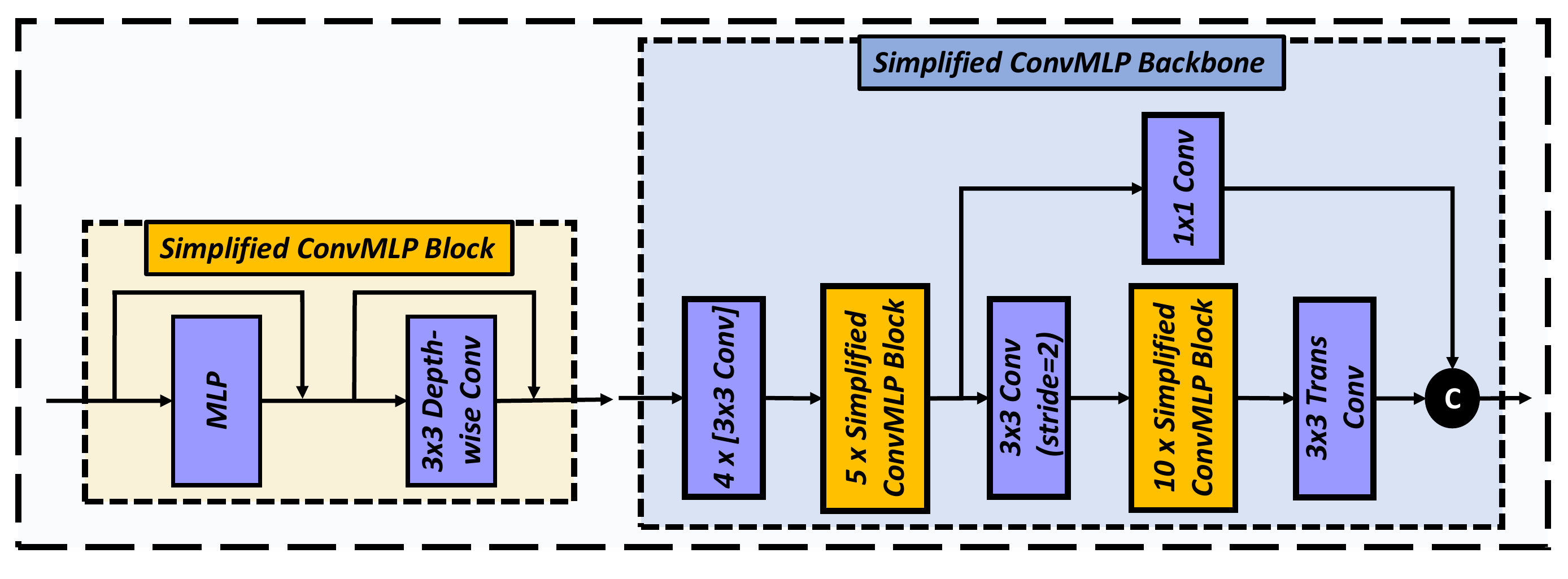}
    \caption{The proposed 2D backbone for the detection task.}
    \label{fig:2dbackbone}
\end{figure}

Inspired by the ConvMLP \cite{li2021convmlp} used in image domains, we propose a simplified version of this architecture to process the BEV-projected features from the 3D backbone before feeding them to the detection head. The simplified ConvMLP (SC) block and the overall proposed 2D backbone architecture are shown in Figure \ref{fig:2dbackbone}. Compared to the original ConvMLP block, we remove the last MLP layer and add a skip connection over the convolution layer to further ease the gradient flow.
In this architecture, the MLP block enables the interaction of features 
in each spatial location, while the subsequent depth-wise convolution
enables efficient space-wise interaction. 
In the backbone,
consecutive Conv blocks (each consists of a convolution layer followed by batch-normalization and ReLU) are first applied to enhance features interactions spatial-wise. Then, resulting features are sent through the first set of SC blocks, 
down-sampled, and fed to another set of SC blocks. The outputs of these two sets of SC blocks are then matched and concatenated as the final set of the 2D features, which is sent to the detection head. 

Compared to the regular 2D backbone in \cite{lang2019pointpillars, yin2021center}, the proposed 2D backbone boosts the detection performance without a steep increase in the model complexity. More specifically, compared to a regular 3x3 convolution layer, an SC block requires 54.6\% less memory and 54.8\% fewer FLOPs. Thus, by replacing regular convolutions with the lighter SC block, we afford to build more consecutive convolutions in a single resolution, achieving a larger receptive field without the need for further down-sampling. In addition, unlike other CNNs that employ a single 1x1 convolution layer for channel depth adjustment, this architecture employs MLP blocks extensively to emphasize on feature extraction within each BEV plane location.

\begin{figure}[t]
    \centering
    \includegraphics[width=1\linewidth]{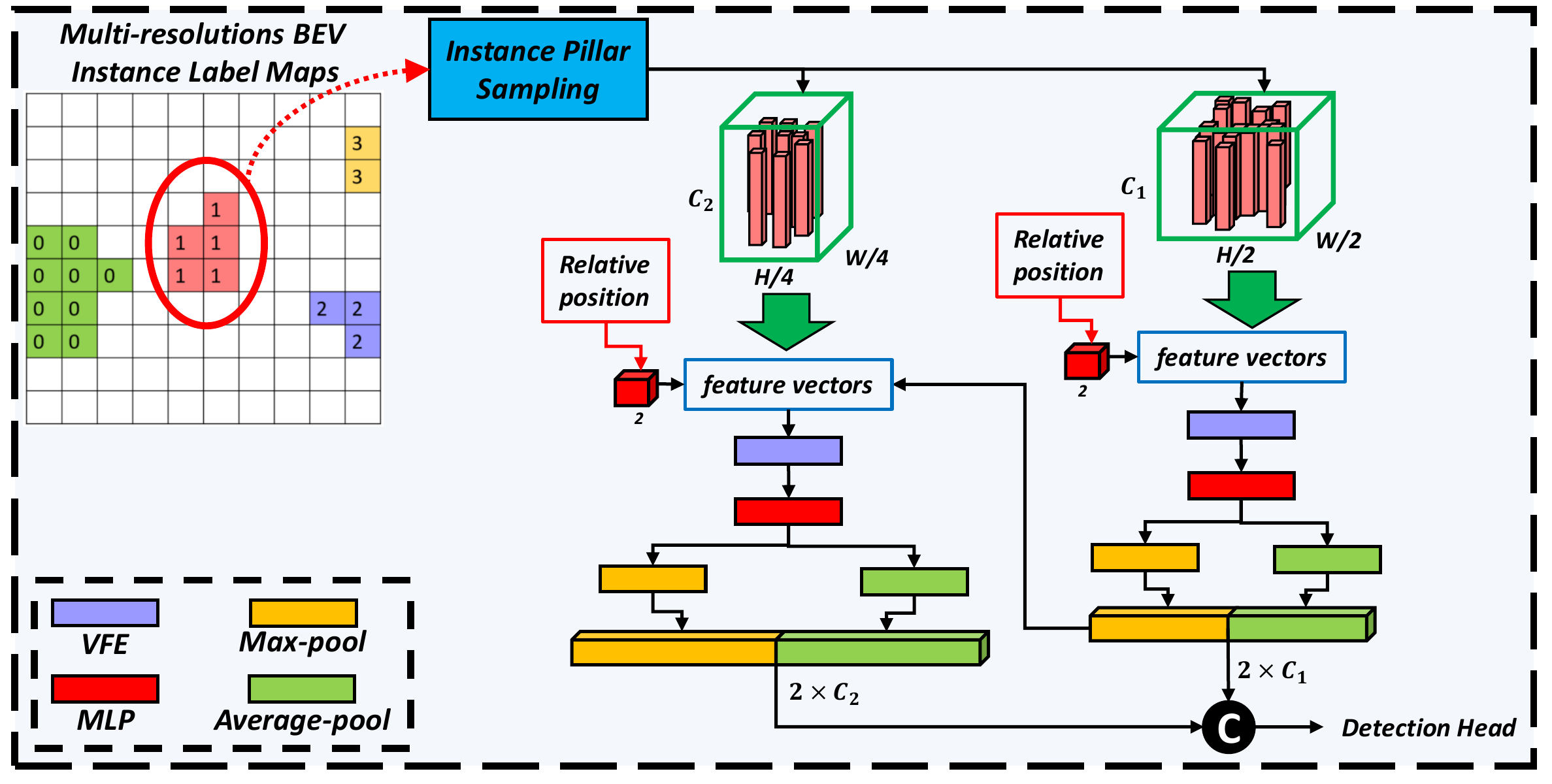}
    \caption{The proposed Instance-based Feature Retrieval (IFR) module. Best viewed in color.}
    \label{fig:ifr}
\end{figure}

\subsection{Instance-based Feature Retrieval (IFR)}

To augment the coarse-scale features extracted by the SC backbone, discarded features during down-sampling operations in the dual-task 3D backbone can be effectively leveraged. For this aim, the IFR module is proposed, shown in Figure \ref{fig:ifr}. This module recovers multi-scale detailed features for each candidate object from $(\frac{Z}{2} \times \frac{H}{2} \times \frac{W}{2})$ and $(\frac{Z}{4} \times \frac{H}{4} \times \frac{W}{4})$ resolutions feature maps in the dual-task 3D backbone. Then, it constructs a new set of features to augment the detection head. 

First,
to reduce computational complexity, on all BEV plane locations, voxel features along the height dimension are averaged to form averaged-voxels features.
Then, a selection strategy is proposed to select averaged-voxels based on instance masks estimated by the panoptic head.
Specifically, given the $l$th scale $s_l$ averaged-voxels features and instance masks of the same scale on the BEV plane, the mean $X$ and $Y$ coordinates of each instance 
are calculated. This gives the mass center location for each instance.
Then, from all the BEV locations that represent each instance, the $K_{s_l}$ nearest averaged-voxels to each instance mass center are selected.

After sampling $K_{s_l}$ averaged-voxels for each instance, the relative coordinates of each sampled averaged-voxel to its instance mass center on both $x-$ and $y-$axis are computed and concatenated to the corresponding feature vector as relative position embedding. This allows the IFR module to be aware of the geometry of sampled averaged-voxels for each instance.
These feature vectors go through a VFE \cite{zhou2018voxelnet} and an MLP layer consecutively. Then, the resulting feature vectors for each instance are pooled using max- and average-pooling layers and concatenated. This is illustrated in the following equations:

\begin{equation}
    v_{j,s_{l}}^{i} = MLP(VFE(Concat(f_{j,s_{l}}^{i}, p_{j,s_{l}}^{i})))
\end{equation}
\begin{equation}
    v_{s_{l}}^{i} = Concat(AvgPool(v_{j,s_{l}}^{i}), MaxPool(v_{j,s_{l}}^{i}))
\end{equation}
where $f_{j,s_{l}}^{i}$ and $p_{j,s_{l}}^{i}$ denote the feature vector and position embedding vector for the $j$th averaged-voxel belonging to $i$th instance in $l$th scale, respectively.

Each resulting single feature vector $v_{s_{l}}^{i}$ encodes and summarizes the sampled averaged-voxels features of the $i$th instance that it corresponds to. The extracted features of an instance in the higher resolution $s_l$ are concatenated to every sampled averaged-voxel feature vector of that instance in the lower resolution $s_{l+1}$ using a cascade connection prior to feeding to the VFE layer. This enables the lower resolution averaged-voxels of an instance to leverage the higher resolution encoded features of the same instance. Finally, the resulting encoded feature vectors of each instance in different resolutions are concatenated and distributed to all the BEV locations that correspond to the instance according to the coarse-scale instance masks. This new set of feature maps is then concatenated to the output features from the 2D backbone and fed to the detection head. By doing so, we effectively augment the detection head by recovering and processing multi-scale information that is unique for each instance and commonly lost prior to the 2D backbone. 


\section{Experiments}

\subsection{Implementation Details}
\label{subsec:impl}
The proposed framework is implemented using the PyTorch  \cite{paszke2019pytorch} and OpenPCDet \cite{openpcdet2020} libraries.
AOP-Net is based on the single-stage CenterPoint detection method. 
For panoptic segmentation, we received the original CPSeg source code \cite{li2021cpseg} from the authors. 
The network was trained from scratch for 140 epochs with Adam optimizer on 8 Tesla V100 GPUs. The One Cycle policy was used for learning rate scheduling with an initial rate of $10^{-3}$. Also, the weight decay was set to $10^{-2}$. In IFR module, we used 2 mid- and high-resolution feature maps from the dual-task 3D backbone and set the $K_{s_1}$ to 16 and $K_{s_2}$ to 25. $c_{1}$, $c_{2}$, $H$, $W$, and $Z$ are set to be 32, 64, 1024, 1024, and 32, respectively. The hidden ratio for MLP in the SC block, IFR's VFE, and IFR's MLP are set to be 2, 4, and 4, respectively.


\subsection{Dataset}
\label{subsec:dataset}
nuScenes \cite{caesar2020nuscenes} 
is a large-scale dataset for autonomous driving that includes both 3D object detection and panoptic segmentation labels.
For 3D object detection, mean Average Precision (mAP) is a metric that is used for evaluation on this benchmark. Moreover, 
nuScenes Detection Score (NDS) is another metric used, which is a weighted sum of mAP and box estimation quality metrics that account for translation, scale, orientation, attributes, and velocity. For 3D panoptic segmentation, we use the mean Panoptic Quality (PQ), which considers both mean Recognition Quality (RQ) and mean Segmentation Quality (SQ), to evaluate the performance.

Waymo Open Dataset \cite{sun2020scalability} is a large-scale 3D object detection dataset. As it lacks panoptic segmentation labels, we prepared the instance and foreground semantic labels using ground truth 3D bounding boxes, and assigned a single background class to all points outside bounding boxes. We report the mAP and the mean Average Precision weighted by Heading (mAPH) for the 3D object detection task. For Waymo, we trained the proposed model on $20\%$ of training data and evaluated on the whole validation data.

\subsection{Results}
\label{subsec:results}

\begin{table*}[t]
\centering
\caption{3D object detection comparison of the AOP-Net and CenterPoint \cite{yin2021center} on nuScenes validation set. CV, Ped, Motor, Bic, and TC are abbreviations for Construction Vehicle, Pedestrian, Motorcycle, Bicycle, and Traffic Cone.}
\label{table:3dodval}
\resizebox{2.1\columnwidth}{!}{
\begin{tabular} {p{2.75cm} p{1cm} p{1cm} p{1cm} p{1cm} p{1cm} p{1.25cm} p{1cm} p{1cm} p{1.25cm} p{1cm} p{1cm} p{1.25cm}} 
\hline\noalign{\smallskip}
\hfil \textbf{Method} & \hfil \textbf{mAP} & \hfil \textbf{NDS} & \hfil \textbf{Car} & \hfil \textbf{Truck} & \hfil \textbf{Bus} & \hfil \textbf{Trailer} & \hfil \textbf{CV} & \hfil \textbf{Ped} & \hfil \textbf{Motor} & \hfil \textbf{Bic} & \hfil \textbf{TC} & \hfil \textbf{Barrier} \\
\noalign{\smallskip}
\hline
\noalign{\smallskip}
\hfil CenterPoint \cite{yin2021center} & \hfil 56.4 & \hfil 64.8 & \hfil 84.7 & \hfil 54.8 & \hfil 67.2 & \hfil 35.3 & \hfil 17.1 & \hfil \textbf{82.9} & \hfil 57.4 & \hfil 35.9 & \hfil 63.3 & \hfil 65.1 \\
\hfil AOP-Net & \hfil \textbf{61.2} & \hfil \textbf{68.5} & \hfil \textbf{85.2} & \hfil \textbf{58.0} & \hfil \textbf{69.4} & \hfil \textbf{42.5} & \hfil \textbf{19.2} & \hfil 82.6 & \hfil \textbf{61.9} & \hfil \textbf{38.9} & \hfil \textbf{72.9} & \hfil \textbf{83.7} \\
\rowcolor{LGreen}
\hfil \textit{Improvement} & \hfil \textit{+4.8} & \hfil \textit{+3.7} & \hfil \textit{+0.5} & \hfil \textit{+3.2} & \hfil \textit{+2.2} & \hfil \textit{+7.2} & \hfil \textit{+2.1} & \hfil \textit{-0.3} & \hfil \textit{+4.5} & \hfil \textit{+3.0} & \hfil \textit{+9.6} & \hfil \textit{+18.6} \\
\hline
\end{tabular}
}
\end{table*}

\begin{table}[t]
\centering
\caption{3D object detection comparison of the AOP-Net and CenterPoint \cite{yin2021center} on Waymo validation set (mAP/mAPH)}
\label{table:wd}
\resizebox{1\columnwidth}{!}{
\begin{tabular} {p{2cm} p{1.2cm} p{1.2cm} p{1.2cm} p{1.2cm} p{1.2cm} p{1.2cm}} 
\hline\noalign{\smallskip}
\hfil \textbf{Method} & \hfil \textbf{Car L1} & \hfil \textbf{Car L2} & \hfil \textbf{Ped L1} & \hfil \textbf{Ped L2} & \hfil \textbf{Cyc L1} & \hfil \textbf{Cyc L2} \\
\noalign{\smallskip}
\hline
\noalign{\smallskip}
\hfil CenterPoint \cite{yin2021center} & \hfil 71.3/70.8 & \hfil 63.2/62.7 & \hfil 72.1/65.5 & \hfil 64.3/58.2 & \hfil 68.7/67.4 & \hfil 66.1/64.9 \\
\hfil AOP-Net & \hfil \textbf{73.2/72.6} & \hfil \textbf{65.0/64.5} & \hfil \textbf{73.1/66.4} & \hfil \textbf{65.2/59.2} & \hfil \textbf{71.0/69.8} & \hfil \textbf{68.7/67.5} \\
\rowcolor{LGreen}
\hfil \textit{Improvement} & \hfil \textit{+1.9/+1.8} & \hfil \textit{+1.8/+1.8} & \hfil \textit{+1.0/+0.9} & \hfil \textit{+0.9/+1.0} & \hfil \textit{+2.3/+2.4} & \hfil \textit{+2.6/+2.6} \\
\hline
\end{tabular}
}
\end{table}

\begin{table*}[t]
\centering
\caption{Performance comparison of 3D object detection methods on nuScenes test set.}
\label{table:3dodtest}
\resizebox{2.1\columnwidth}{!}{
\begin{tabular} {p{4cm} p{1.5cm} p{1.5cm} p{1.5cm} p{1.5cm} p{1.5cm} p{1.5cm} p{1.5cm}} 
\hline\noalign{\smallskip}
\hfil \textbf{Method} & \hfil \textbf{mAP} $\uparrow$  & \hfil \textbf{mATE} $\downarrow$ & \hfil \textbf{mASE} $\downarrow$ & \hfil \textbf{mAOE} $\downarrow$ & \hfil \textbf{mAVE} $\downarrow$ & \hfil \textbf{mAAE} $\downarrow$ & \hfil \textbf{NDS} $\uparrow$ \\
\noalign{\smallskip}
\hline
\noalign{\smallskip}
\hfil PointPillars \cite{cvpr} & \hfil 30.5 & \hfil 51.7 & \hfil 29.0 & \hfil 50.0 & \hfil 31.6 & \hfil 36.8 & \hfil 45.3 \\
\rowcolor{Gray}
\hfil CBGS \cite{zhu2019class} & \hfil 52.8 & \hfil 30.0 & \hfil 24.7 & \hfil 37.9 & \hfil 24.5 & \hfil 14.0 & \hfil 63.3 \\
\hfil CVCNet \cite{zhu2021cross} & \hfil 55.3 & \hfil 30.0 & \hfil 24.4 & \hfil 38.9 & \hfil 26.8 & \hfil \textbf{12.2} & \hfil 64.4 \\
\rowcolor{Gray}
\hfil HotSpotNet \cite{chen2020object} & \hfil 59.3 & \hfil 27.4 & \hfil 23.9 & \hfil 38.4 & \hfil 33.3 & \hfil 13.3 & \hfil 66.0 \\
\hfil Multi-task \cite{cvpr} & \hfil \textbf{60.9} & \hfil 28.8 & \hfil 24.5 & \hfil 40.0 & \hfil 25.3 & \hfil 12.8 & \hfil 67.3 \\
\rowcolor{Gray}
\hfil AOP-Net & \hfil 60.6 & \hfil \textbf{28.0} & \hfil \textbf{24.2} & \hfil \textbf{36.2} & \hfil \textbf{22.1} & \hfil \textbf{12.2} & \hfil \textbf{68.1} \\
\hline
\end{tabular}}
\end{table*}


\subsubsection{3D Object Detection} In Table. \ref{table:3dodval} and \ref{table:wd}, we compare the evaluation results between the proposed method and CenterPoint on the nuscenes and Waymo validation sets. The AOP-Net is based on the CenterPoint first stage. As shown, the proposed method outperforms the CenterPoint in both mAP and NDS scores for nuScenes significantly, and mAP and mAPH for Waymo considerably. As elaborated in ablations, improvements in the detection of large and small objects can be attributed to the SC Backbone and the IFR module, respectively.

The comparison between AOP-Net and other published state-of-the-art 3D object detection methods on the nuScenes test set are shown in Table \ref{table:3dodtest}. It can be seen that the proposed method outperforms all other methods in terms of NDS and all five error metrics that represent the box estimation quality, including the mean average errors in translation (mATE), scale (mASE), orientation (mAOE), velocity (mAVE), and attribute (mAAE). This improvement can be attributed to the guidance received from the panoptic segmentation module, 
both direct (exploitation of panoptic segmentation predictions in IFR) and indirect (back propagation of panoptic loss in backbones).

\begin{figure}[t]
    \centering
    \includegraphics[width=1\linewidth]{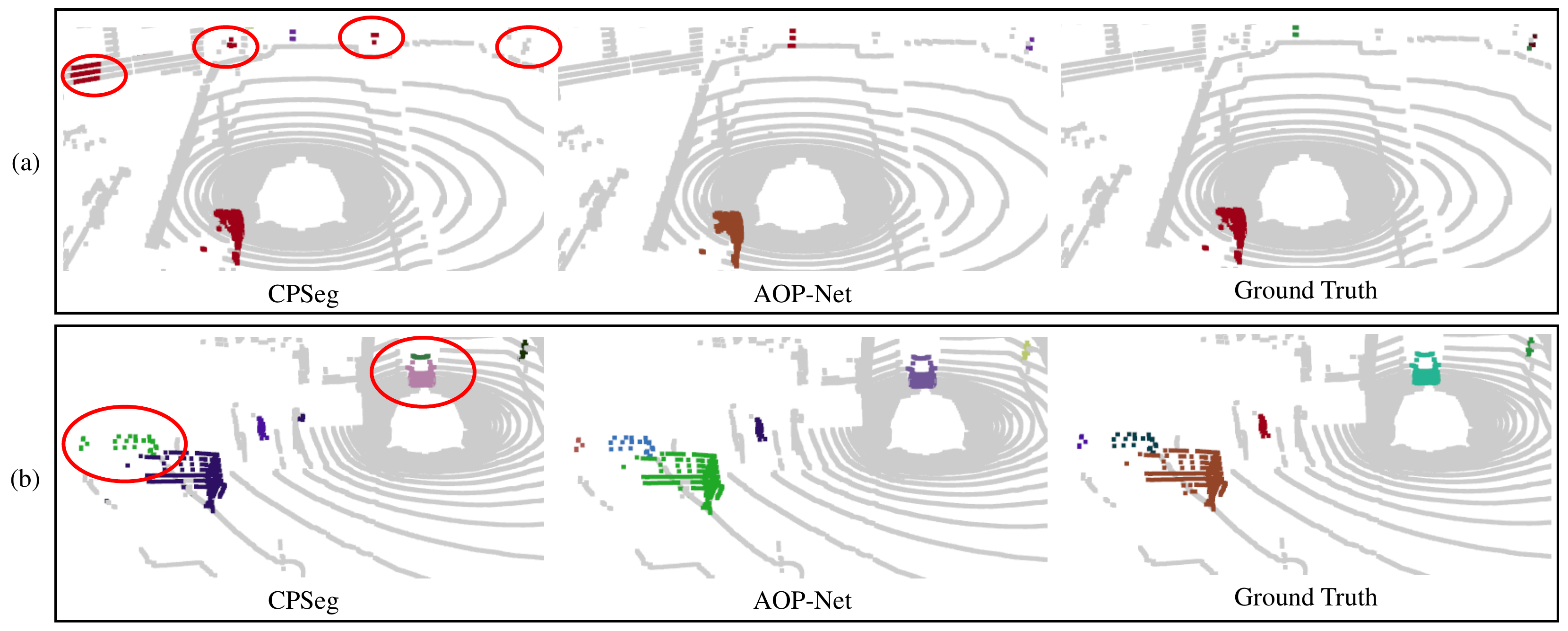}
    \caption{Comparison of instance segmentation results between CPSeg and AOP-Net. Best viewed in color.}
    \label{fig:3dpsviz}
\end{figure}

\subsubsection{3D Panoptic Segmentation}

In Table \ref{table:3dpstest}, comparing AOP-Net with other state-of-the-art published methods on the nuScenes test set, we validate that the AOP-Net obtains higher mean PQ. Compared to the second row, which is a standalone simplified version of CPSeg originally incorporated in AOP-Net, the AOP-Net receives the additional injection of multi-scale detection-level features, which lead to significantly better panoptic performance.

\begin{table*}[t]
\centering
\caption{Performance comparison of 3D panoptic segmentation methods on nuScenes test set.}
\label{table:3dpstest}
\resizebox{2.1\columnwidth}{!}{
\begin{tabular} {p{3.75cm} p{1cm} p{1cm} p{1cm} p{1cm} p{1cm} p{1cm} p{1cm} p{1cm} p{1cm} p{1.25cm}}
\hline\noalign{\smallskip}
\hfil \textbf{Method} & \hfil \textbf{PQ} & \hfil \textbf{RQ} & \hfil \textbf{SQ} & \hfil \textbf{PQ\textsuperscript{Th}} & \hfil \textbf{RQ\textsuperscript{Th}} & \hfil \textbf{SQ\textsuperscript{Th}} & \hfil \textbf{PQ\textsuperscript{St}} & \hfil \textbf{RQ\textsuperscript{St}} & \hfil \textbf{SQ\textsuperscript{St}} & \hfil \textbf{mIOU} \\
\noalign{\smallskip}
\hline
\noalign{\smallskip}
\hfil PanopticTrackNet \cite{hurtado2020mopt} & \hfil 51.6 & \hfil 63.3 & \hfil 80.4 & \hfil 45.9 & \hfil 56.1 & \hfil 81.4 & \hfil 61.0 & \hfil 75.4 & \hfil 79.0 & \hfil 58.9 \\
\rowcolor{Gray}
\hfil AOP-Net (Single-task) & \hfil 62.1 & \hfil 72.0 & \hfil 85.8 & \hfil 59.3 & \hfil 66.9 & \hfil 87.9 & \hfil 66.8 & \hfil 80.5 & \hfil 82.2 & \hfil 67.8 \\
\hfil EfficientLPS \cite{sirohi2021efficientlps} & \hfil 62.4 & \hfil 74.1 & \hfil 83.7 & \hfil 57.2 & \hfil 68.2 & \hfil 83.6 & \hfil 71.1 & \hfil \textbf{84.0} & \hfil 83.8 & \hfil 66.7 \\
\rowcolor{Gray}
\hfil Panoptic-PolarNet \cite{zhou2021panoptic} & \hfil 63.6 & \hfil 75.1 & \hfil 84.3 & \hfil 59.0 & \hfil 69.8 & \hfil 84.3 & \hfil \textbf{71.3} & \hfil 83.9 & \hfil \textbf{84.2} & \hfil 67.0 \\
\hfil AOP-Net & \hfil \textbf{68.3} & \hfil \textbf{78.2} & \hfil \textbf{86.9} & \hfil \textbf{67.3} & \hfil \textbf{75.6} & \hfil \textbf{88.6} & \hfil 69.8 & \hfil 82.6 & \hfil 84.0 & \hfil \textbf{72.5} \\
\hline
\end{tabular}}
\end{table*}

\begin{table}[t]
\centering
\caption{Effect of individual components on detection performance on nuScenes validation set}
\label{table:components}
\resizebox{1\columnwidth}{!}{
\begin{tabular} {p{4cm} p{4cm} p{1cm} p{1cm} p{1cm}} 
\hline\noalign{\smallskip}
\hfil \textbf{Dual-task 3D backbone} & \hfil \textbf{Simplified ConvMLP} & \hfil \textbf{IFR} & \hfil \textbf{mAP} & \hfil \textbf{NDS} \\
\noalign{\smallskip}
\hline
\noalign{\smallskip}
\hfil  & \hfil  & \hfil  & \hfil 56.9 & \hfil 65.4 \\
\rowcolor{Gray}
\hfil  & \hfil  \checkmark & \hfil \checkmark & \hfil 58.9 & \hfil 67.1 \\
\hfil \checkmark & \hfil  & \hfil & \hfil 59.8  & \hfil  66.9 \\
\rowcolor{Gray}
\hfil \checkmark & \hfil \checkmark & \hfil  & \hfil 60.7 & \hfil 68.0 \\
\hfil \checkmark & \hfil \checkmark & \hfil \checkmark & \hfil \textbf{61.4} & \hfil \textbf{68.5} \\
\hline
\end{tabular}}
\end{table}

\begin{table}[t]
\centering
\caption{Effect of 3D backbone on detection and panoptic segmentation performance on nuScenes validation set}
\label{table:backbone}
\resizebox{1\columnwidth}{!}{
\begin{tabular} {p{4cm} | p{1cm} p{1cm} | p{1cm} p{1cm} p{1cm}  p{1.25cm}} 
\hline\noalign{\smallskip}
\hfil \textbf{Module} & \hfil \textbf{mAP} & \hfil \textbf{NDS} & \hfil \textbf{PQ} & \hfil \textbf{RQ} & \hfil \textbf{SQ} & \hfil \textbf{mIOU} \\
\noalign{\smallskip}
\hline
\noalign{\smallskip}
\hfil 3D Backbone & \hfil 58.9 & \hfil 67.1 & \hfil 72.6 & \hfil 83.1 & \hfil 86.9 & \hfil 72.4 \\
\rowcolor{Gray}
\hfil Dual-task 3D Backbone & \hfil \textbf{61.2} & \hfil \textbf{68.5} & \hfil \textbf{75.6} & \hfil \textbf{85.9} & \hfil \textbf{87.7} & \hfil \textbf{75.9} \\
\hline
\end{tabular}}
\end{table}

\begin{table}[t]
\centering
\caption{Effect of 2D backbone on detection of large objects on nuScenes validation set.}
\label{table:rev1_1}
\resizebox{1\columnwidth}{!}{
\begin{tabular} {p{3cm} p{1cm} p{1cm}  p{1cm} p{1cm}} 
\hline\noalign{\smallskip}
\hfil \textbf{Module} & \hfil \textbf{Truck} & \hfil \textbf{Bus} & \hfil \textbf{Trailer} & \hfil \textbf{CV} \\
\noalign{\smallskip}
\hline
\noalign{\smallskip}
\hfil Traditional 2D Backbone & \hfil 56.6 & \hfil 67.4 & \hfil 41.1 & \hfil \textbf{21.4} \\
\rowcolor{Gray}
\hfil SC Backbone & \hfil \textbf{57.4} & \hfil \textbf{68.8} & \hfil \textbf{42.3} & \hfil 20.5 \\
\hline
\end{tabular}}
\end{table}

\begin{table}[t]
\centering
\caption{Effect of IFR module on detection of small objects on nuScenes validation set.}
\label{table:rev1_2}
\resizebox{1\columnwidth}{!}{
\begin{tabular} {p{1.5cm} p{1cm} p{1cm}  p{1cm} p{1cm} p{1cm}} 
\hline\noalign{\smallskip}
\hfil \textbf{Module} & \hfil \textbf{Ped} &  \hfil \textbf{Motor} & \hfil \textbf{Bic} & \hfil \textbf{TC} & \hfil \textbf{Barrier} \\
\noalign{\smallskip}
\hline
\noalign{\smallskip}
\hfil Without IFR & \hfil 81.4 & \hfil 59.7 & \hfil 36.9 & \hfil 72.7 & \hfil 82.6 \\
\rowcolor{Gray}
\hfil With IFR & \hfil \textbf{82.6} & \hfil \textbf{61.9} & \hfil \textbf{38.9} & \hfil \textbf{72.9} & \hfil \textbf{83.7} \\
\hline
\end{tabular}}
\end{table}

\begin{table}[t]
\centering
\caption{Detection performance comparison of ConvMLP backbones on nuScenes validation set.}
\label{table:rev1_0}
\resizebox{1\columnwidth}{!}{
\begin{tabular} {p{3cm} p{2.3cm} p{2cm}  p{1cm} p{1cm}} 
\hline\noalign{\smallskip}
\hfil \textbf{Module} & \hfil \textbf{ConvMLP Blocks} & \hfil \textbf{\# Params (M)} & \hfil \textbf{mAP} & \hfil \textbf{NDS} \\
\noalign{\smallskip}
\hline
\noalign{\smallskip}
\hfil Original ConvMLP & \hfil 2, 6 & \hfil 2.4 & \hfil 61.1 & \hfil 67.9 \\
\rowcolor{Gray}
\hfil Simplified ConvMLP & \hfil 2, 5 & \hfil 1.8 & \hfil 59.9 & \hfil 67.2 \\
\hfil Simplified ConvMLP & \hfil 5, 10 & \hfil 2.4 & \hfil \textbf{61.2} & \hfil \textbf{68.5} \\
\rowcolor{Gray}
\hfil Simplified ConvMLP & \hfil 10, 20 & \hfil 3.4 & \hfil 60.7 & \hfil 67.8 \\
\hline
\end{tabular}}
\end{table}

In Figure \ref{fig:3dpsviz}, the benefits of the unified multi-task framework towards panoptic segmentation are visible. In example (a), the standalone CPSeg struggles to predict the semantics of distant points, leading to three false positives and one false negative. In (b), CPSeg under-segments on the left and over-segments near the top as it is less confident about regions that are less visible behind a large body of points.
In both cases, the dual-task 3D backbone in the AOP-Net provides effective multi-scale 3D features to prevent these errors. 

\begin{table}[t]
\centering
\caption{Performance of other BEV-based 3D object detection methods in AOP-Net on nuScenes validation set.}
\label{table:otherdetectors}
\resizebox{1\columnwidth}{!}{
\begin{tabular} {p{4cm} p{3cm} p{2cm} p{2cm}} 
\hline\noalign{\smallskip}
\hfil \textbf{Method} & \hfil \textbf{\# Params (M)} & \hfil \textbf{mAP} & \hfil \textbf{NDS} \\
\noalign{\smallskip}
\hline
\noalign{\smallskip}
\hfil PointPillars \cite{lang2019pointpillars} & \hfil 6.1 & \hfil 44.6 & \hfil 58.1 \\
\hfil Complex PointPillars & \hfil 13.7 & \hfil 44.3  & \hfil 57.7 \\
\hfil AOP-Net(PointPillars) & \hfil 13.0 & \hfil \textbf{54.5} & \hfil \textbf{64.0} \\
\rowcolor{LGreen}
\hfil \textit{Improvement} & \hfil - & \hfil \textit{+9.9} & \hfil \textit{+5.9} \\
\hline
\hfil SECOND \cite{yan2018second} & \hfil 9.0 & \hfil 51.8 & \hfil 62.7 \\
\hfil Complex SECOND & \hfil 13.9 & \hfil 52.1 & \hfil 62.8 \\
\hfil AOP-Net(SECOND) & \hfil 14.6 & \hfil \textbf{58.2} & \hfil \textbf{65.7} \\
\rowcolor{LGreen}
\hfil \textit{Improvement} & \hfil - & \hfil \textit{+6.1} & \hfil \textit{+2.9} \\
\hline
\end{tabular}}
\end{table}

\begin{figure}[t]
    \centering
    \includegraphics[width=1\linewidth]{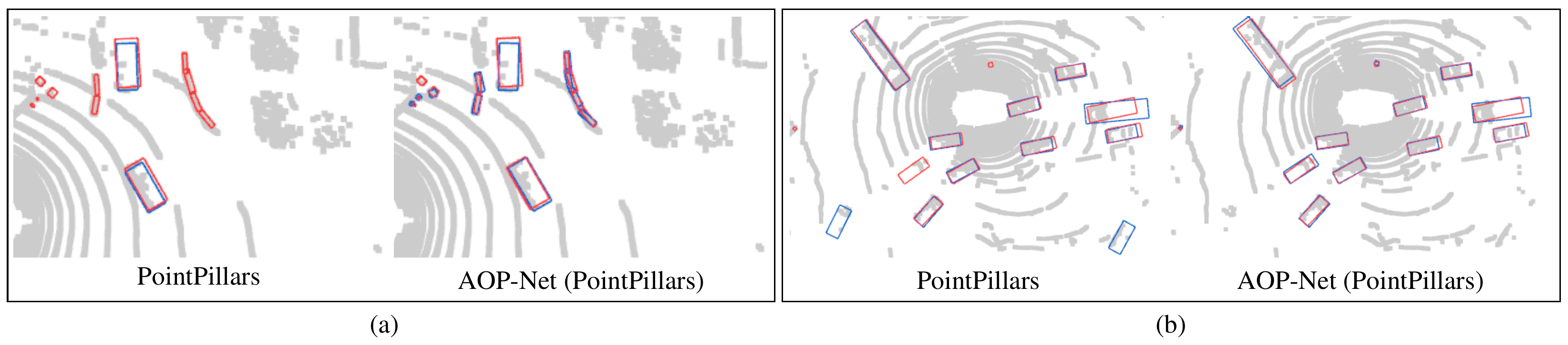}
    \caption{Comparison of qualitative results between PointPillars and AOP-Net (PointPillars) for 3D object detection. The \textcolor{red}{red} and \textcolor{blue}{blue} colors show the ground-truth and the predicted boxes, respectively. Best viewed in color.}
    \label{fig:3dodcomparison}
\end{figure}

\subsection{Ablation Studies}
\label{subsec:ablation}
\subsubsection{Effect of each proposed component} The contributions of AOP-Net modules are shown in Table \ref{table:components}. It can be seen that each and a combination of these modules adapt well to the baseline and provide strong performance gains. 

Specifically, in Table \ref{table:backbone}, it can be seen that incorporating the dual-task 3D backbone significantly boosts performances for both tasks. In particular, the improvement of AOP-Net in panoptic segmentation is mainly attributed to this module. As the 3D backbone is conditioned on both tasks, the learned features are enriched and provide additional clues regarding foreground objects. Moreover, the 3D backbone captures features without the occlusion or scale-variant issues common for feature extraction in RV plane. When projected to RV and fused with already extracted RV-based features, these set of features are more reliable and helpful in segmenting occluded and distant objects. These factors lead to a significant improvement in both mIOU and PQ. 


In Table \ref{table:rev1_1}, we demonstrate that improvements in the detection of large class objects can be attributed to the enlarged receptive fields and more extensive channel-wise feature extraction from the SC Backbone. 

In Table \ref{table:rev1_2}, it can be seen that IFR plays a strong role in better detecting small isolated objects. This is because IFR influences the detection head to pay more attention to 
multi-scale features that are relevant to foreground objects. By reintroducing this information that is otherwise lost in the down-sampling process in the 3D backbone, the detection head improves both precision (by refining possible candidates) and recall (retrieving missed objects that are better detected in RV panoptic segmentation).

\subsubsection{Variations of ConvMLP Backbones} 
In Table \ref{table:rev1_0}, a similarly sized network (in terms of \# parameters) that uses original ConvMLPs has fewer consecutive layers and lower performance. Also, comparing rows 2-4, having $5$ and $10$ SC blocks gives the best trade-off in terms of performance and complexity.

\subsubsection{Other BEV-based 3D object detectors in the proposed framework} To show that AOP-Net can also work with anchor-based detection methods, 
we performed experiments by adapting the AOP-Net to PointPillars \cite{lang2019pointpillars} and SECOND \cite{yan2018second}. The results of these experiments are shown in Table \ref{table:otherdetectors}. Also, we increased the model complexity of the PointPillars and SECOND and named them as Complex PointPillars and Complex SECOND. It can be seen that by simply increasing the model complexity, the performance boost is either nonexistent or limited. However, under the proposed framework, the mAP and NDS are improved remarkably. 
The effects of the proposed framework are prevalent in Figure \ref{fig:3dodcomparison}. It can be seen that in both examples (a) and (b), due to the loss of fine-scale features during down-sampling, PointPillars fails to detect small objects. On the other hand, in the proposed method, these objects are recognized by the RV-based segmentation module and their fine-scale features are recovered by the IFR module, allowing for their detection. Moreover, in example (b), PointPillars produces two false positives from afar, while the AOP-Net is properly guided by panoptic-level information and circumvents these mistakes. 

\section{Conclusions}
We propose AOP-Net, an all-in-one perception framework for LiDAR-based joint 3D object detection and panoptic segmentation. In this framework, 
we design the dual-task 3D backbone to consider both semantic- and instance-level information of the scene, thereby augmenting both the BEV-based detection head and RV-based panoptic head.
Also, the multi-scale 3D voxel features resulted from this backbone are used to augment the single-scale RV feature maps in the panoptic segmentation task. Moreover, a deep and efficient 2D backbone based on the simplified ConvMLP (SC) block is proposed, which results in detection improvement. Finally, to recover features lost during down-sampling operations in the dual-task 3D backbone, a novel instance-based feature retrieval (IFR) module is proposed that relies on predicted instance masks and recovers features to augment the detection backbone. Experimental results on nuScenes and Waymo datasets show strong improvements in both 3D panoptic segmentation and object detection tasks under the proposed framework, while demonstrating that the detection accuracy of any BEV-based 3D object detection can be improved using the proposed strategy.

\end{document}